\documentclass[letterpaper, 10 pt, conference]{ieeeconf}  %

\IEEEoverridecommandlockouts                              %

\usepackage{graphics} %
\usepackage{amsmath} %
\usepackage{amssymb}  %
\usepackage{todonotes}
\usepackage{graphicx}
\usepackage{subcaption}
\usepackage{dblfloatfix}
\usepackage{siunitx}
\usepackage{caption}   %
\usepackage{cuted}     %
\usepackage{textcomp}

\newif\ifanonymized \anonymizedfalse   %

\title{\LARGE \bf
On the Utility of Pre-Trained Visual Representations for Policy Learning of Contact-Rich Insertion Tasks }

\ifanonymized
\author{Anonymous Authors}
\else
\author{Leon Kim$^1$, Dinesh Jayaraman$^{1,\dagger}$, Michael Posa$^{1,\dagger}$
    \thanks{$\dagger$Contributed equally to this work} \thanks{$^{1}$Leon Kim, Dinesh Jayaraman, and Michael Posa are with the GRASP Laboratory, University of Pennsylvania, USA \{leonmkim, dineshj, posa\}@engineering.upenn.edu} }
\fi

\begin{document}
\bstctlcite{IEEEexample:BSTcontrol}

\maketitle

\thispagestyle{empty} \pagestyle{empty}

\begin{figure*}[!b] 
  \centering
  \begin{subfigure}[t]{0.18\textwidth}
    \includegraphics[width=\linewidth]{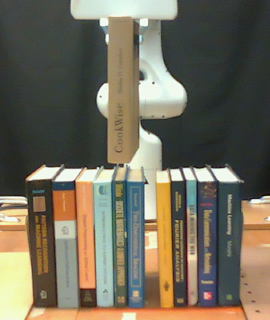}
    \subcaption{Raw RGB frame near beginning of task.}\label{fig:wide-a}
  \end{subfigure}\hfill
  \begin{subfigure}[t]{0.18\textwidth}
    \includegraphics[width=\linewidth]{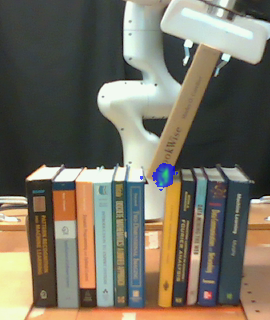}
    \subcaption{Map of estimated contact regions during wedging. Color transitions from blue to green for higher probability.}\label{fig:wide-b}
  \end{subfigure}\hfill
  \begin{subfigure}[t]{0.18\textwidth}
    \includegraphics[width=\linewidth]{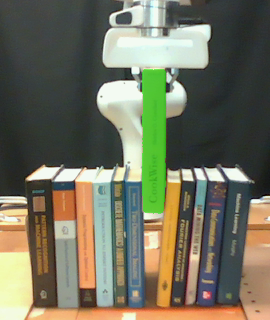}
    \subcaption{Grasped object mask in green partway through insertion.}\label{fig:wide-c}
  \end{subfigure}\hfill
  \begin{subfigure}[t]{0.18\textwidth}
    \includegraphics[width=\linewidth]{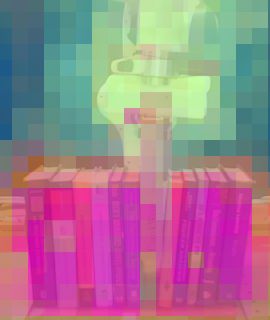}
    \subcaption{PCA of Theia near the end of task.}\label{fig:wide-d}
  \end{subfigure}

  \captionof{figure}{Execution of book-insertion task showcasing different visual representations evaluated in this work. We evaluate policies trained from scratch with raw RGB-D inputs (a), policies augmented with contact-centric features (b), segmentation masks of the grasped object (c), and Theia features (d).}
  \label{fig:teaser}
\end{figure*}

\begin{abstract}
    Visuomotor imitation learning has become a predominant paradigm for real-world robotic manipulation, yet it remains data-inefficient and brittle to task-irrelevant visual shifts. Pre-trained visual representations promise to ease this burden by distilling priors from internet-scale data, but the ``right'' representation for robotics remains unresolved: existing studies span broad but sparse task distributions and reach conclusions that vary and even conflict, pointing to strong task and setting dependence. We examine this question by bringing a magnifying glass to a common contact-rich manipulation task: insertion.  We compare representations from widely used vision foundation models against an explicit contact-centric representation on two closely related insertion tasks, perturb each along controlled axes of occlusion and physical tolerance, and test whether conclusions transfer to the real world. Task dependence proves to be more subtle than previous, broadly scoped evaluations revealed: the best representation differs even between our near-identical tasks, and small changes shift trends in unpredictable ways. Despite this variability, results across simulation and real-world experiments demonstrate the distinct value of spatially-grounded, and in particular, contact-centric information for contact-rich policy learning, while underscoring how difficult it is to estimate such cues reliably under realistic visual complexity and sensor noise. We hope these findings motivate further study of representations in contact-rich manipulation, along with advances in contact perception and its integration into visuomotor policy learning.

\end{abstract}

\section{INTRODUCTION}

Imitation learning has emerged as a predominant paradigm for robotic manipulation, enabling hobbyists and researchers alike to train visuomotor policies for diverse real-world tasks~\cite{chiDiffusionPolicyVisuomotor2023a,chiUniversalManipulationInterface2024b,zhaoLearningFineGrainedBimanual2023a,zhaoALOHAUnleashedSimple2024a}. Yet state-of-the-art imitation learning remains brittle to task-irrelevant visual shifts unless supported by laborious amounts of demonstration data. Pre-trained visual representations promise to ease this burden by leveraging priors distilled from internet-scale datasets~\cite{oquabDINOv2LearningRobust2024,radfordLearningTransferableVisual2021a,zhaiSigmoidLossLanguage2023}. However, which representation is “best” for robotics remains hotly debated: studies to date have evaluated broad but sparse task distributions and reached conclusions that vary and even contradict one another, with most suggesting the answer is highly task- and setting-dependent~\cite{chenRobustPoliciesMidLevel2020c,burnsWhatMakesPreTrained2023a,hansenPreTrainingVisuoMotorControl2023,yangBridgingPerceptionAction2025}. Consistent with this, modern robotics foundation models (RFMs) have not converged on a single recipe, instead adopting disparate pre-trained vision encoders~\cite{teamCarefulExaminationLarge2025a,kimOpenVLAOpenSourceVisionLanguageAction2024,black$p_0$VisionLanguageActionFlow2024}.

This uncertainty is especially salient in fine-grained, contact-rich manipulation. Despite large demonstration datasets and state-of-the-art visual representations, RFMs still struggle with contact-rich behaviors, with performance often lagging behind that on gross motor skills and soft-object manipulation. Perhaps because of this difficulty, fine-grained, contact-rich manipulation remains underrepresented in robot learning: datasets and benchmarks largely emphasize reaching, pick-and-place, or articulated object manipulation where successful behaviors demand neither millimeter-level precision in positioning nor frequent contact mode transitions across different surfaces~\cite{majumdarWhereAreWe2024,jamesRLBenchRobotLearning2019a,yuMetaWorldBenchmarkEvaluation2021a,khazatskyDROIDLargeScaleInTheWild2025,collaborationOpenXEmbodimentRobotic2025,liuLIBEROBenchmarkingKnowledge2023}. When contact-rich tasks do appear, they are often posed with generous tolerances that avoid tight-fit failure modes. 

We argue this gap persists because contact-rich tasks both amplify common challenges and introduce unique ones. Rigid-body contact yields stiff, highly sensitive dynamics where small errors can cause discontinuous outcomes. Tight tolerances and friction induce hard-to-recover failures (e.g., jamming and wedging) that are pervasive in real assemblies~\cite{simunovic1975force,whitneyQuasiStaticAssemblyCompliantly1982a}. Finally, during close proximity and contact, task-relevant surfaces and their gaps are frequently occluded, limiting visual feedback precisely when and where it is most needed. These challenges have motivated efforts to incorporate spatially-dense contact and proximity information of critical object surfaces from tactile (and sometimes visual) sensors into visuomotor policy learning~\cite{bahlAffordancesHumanVideos2023b,guzeySeeTouchLearning2024,pattabiramanLearningPreciseContactRich2024,higueraTactilePixelsMultisensory2025,huang3DViTacLearningFineGrained2025a}, but its value relative to standard visual priors remains unclear.

In this work, we restrict the question of the ``right'' visual representation to contact-rich manipulation, specifically insertion, and ask: (1) Do general-purpose vision foundation model (VFM) features (e.g., DINOv2~\cite{oquabDINOv2LearningRobust2024}, SAM-2~\cite{raviSAM2Segment2024}, Theia~\cite{shangTheiaDistillingDiverse2024}) improve policy performance over training from scratch? (2) Can explicit contact-centric information outperform VFMs, and under what conditions? Prior studies have pursued such questions over broad sets of disparate tasks, drawing preliminary conclusions that contact-rich tasks will benefit more from contact representations~\cite{bahlAffordancesHumanVideos2023b,huang3DViTacLearningFineGrained2025a}.
These studies are beneficial for surveying the landscape of manipulation tasks, but we argue here that this breadth obscures the severity of the very task-dependence they report. Here, we specifically study a narrow domain, two similar insertion tasks, where previous studies would have reasonably indicated that contact features would bring value to both tasks. By evaluating policies in a narrow task domain, we are able to identify and highlight the surprising sensitivity in our findings to subtle changes in these tasks. To expose setting-dependent sensitivity, we perturb tasks along controlled axes: camera viewpoint shifts that modulate occlusion, and the tightness of physical tolerances. Finally, we extend our evaluation to the real world, where visual complexity and sensor noise provide further nuance to simulation results.

As this manuscript will detail, our results identify a surprising level of sensitivity, often defying traditional expectations. (1) In keeping with published results, we fully expected VFMs, particularly the robotics-tailored Theia, to substantially improve policy performance. However, while at least one VFM managed to provide substantial gains in most settings, the best-performing representation differed across tasks: Theia performed best in our real-world task but consistently matched baselines across simulated ones. Meanwhile, segmentation masks performed best in one simulated task but degraded once we introduced occlusions.
(2) Regarding contact-specific features, we hypothesized that explicit contact information would be more useful than common VFMs. Ground-truth (GT) contact information indeed consistently outperformed other representations across all settings, pointing to the critical importance of directly collocated tactile-like spatial features. However, when occlusions were minimal, segmentation masks performed just as well. Furthermore, estimated contact features struggled to match the performance of their ground-truth counterparts when faced with heavy occlusions or real-world sensor noise.

In sum, our contributions are as follows:
\begin{enumerate}
    \item We take a hotly contested question in robot learning, restrict it to fine-grained, contact-rich insertion tasks, and demonstrate that the benefit of pre-trained visual representations is even less predictable than previously suggested.
    \item Our results highlight not only that the best representation is highly task-dependent, but that seemingly small variations in a task can affect trends in unpredictable ways. 
    \item Despite this variability, our study provides evidence that contact information can offer distinct value, while also underscoring the practical challenge of inferring it reliably under occlusion and sensor noise.
\end{enumerate}
Ultimately, our findings show that within the contact-rich domain, it is difficult to anticipate the right representation among common VFMs. While explicitly incorporating contact information into visuomotor policy learning holds promise, realizing that promise remains challenging. We offer this study less as a prescription than as a controlled demonstration that no reliable prescription currently exists. This motivates further research into the role of visual representations in contact-rich settings, the advancement of robust contact perception capabilities, and their effective integration into policy learning algorithms.

\section{RELATED WORK}

Leveraging pre-trained visual representations for manipulation is an active topic in robot learning. Chen et al.~\cite{chenRobustPoliciesMidLevel2020c} studied interpretable mid-level visual representations and found surface normals and “3D edges” provide consistent gains across simulated reaching, pushing, and pick-and-place tasks, though the best representation remains highly task-dependent. 

Subsequent works proposed representations tailored for robotics, including R3M~\cite{nairR3MUniversalVisual2022}, MVP~\cite{xiaoMaskedVisualPretraining2022}, VIP~\cite{maVIPUniversalVisual2023}, PVR~\cite{parisiUnsurprisingEffectivenessPreTrained2022a}, and VC-1~\cite{majumdarWhereAreWe2024}. Recently, Shang et al.~\cite{shangTheiaDistillingDiverse2024} introduced Theia, which distills multiple foundation models to achieve performance gains over general-purpose VFMs (e.g., CLIP, SAM, DINOv2) and robotics-specific models (e.g., R3M, MVP, VC-1).

Concurrent works debate the necessity of specialized pre-training. Burns et al.~\cite{burnsWhatMakesPreTrained2023a} find that effective representations are predictive of object segmentation, suggesting surface saliency helps policies. Conversely, Hansen et al.~\cite{hansenPreTrainingVisuoMotorControl2023} argue a designed from-scratch baseline can match pre-trained features such as PVR~\cite{parisiUnsurprisingEffectivenessPreTrained2022a}, MVP~\cite{xiaoMaskedVisualPretraining2022}, and R3M~\cite{nairR3MUniversalVisual2022}.

Crucially, these studies rely on simulation benchmarks prioritizing gross motor skills over fine-grained tasks. MetaWorld~\cite{yuMetaWorldBenchmarkEvaluation2021a} is a popular choice~\cite{burnsWhatMakesPreTrained2023a,majumdarWhereAreWe2024,shangTheiaDistillingDiverse2024}, wherein out of 50 tasks, only two involve insertion with generous tolerances. Another frequent choice is FrankaKitchen~\cite{guptaRelayPolicyLearning2019}, which consists largely of reaching, pick-and-place, and articulated object manipulation~\cite{burnsWhatMakesPreTrained2023a,parisiUnsurprisingEffectivenessPreTrained2022a}.

Most similar to our work, Yang et al.~\cite{yangBridgingPerceptionAction2025} investigate several ``spatially-grounded'' representations. Notably, they include a simulated peg-insertion and puzzle-board assembly task. Consistent with the literature, they find different tasks benefit from different representations: object position traces are best for peg-insertion while object poses are preferable for assembly. However, no details of object dimensions, tolerances, or episodic randomizations are provided, making it difficult to contextualize the results. In contrast, our work explicitly focuses on tightly-toleranced insertion, featuring 3 mm tolerances for peg-insertion and negative clearance for the book-insertion task, pushing the demand for precise control and contact-rich interactions.

Our work builds on insights from the literature while maintaining a tightly-scoped task regime. We adopt DINOv2, Theia, segmentation masks, and training-from-scratch as baselines in our study, but we explicitly focus on fine-grained, contact-rich manipulation, where we hypothesize success hinges on representing the precise spatial relationships between task-relevant objects. To this end, we introduce a novel contact-centric visual representation specific to our study which is designed to expose gaps and local geometry salient to contact-rich behaviors.

\section{METHODOLOGY}

\begin{figure*}[!b] 
  \centering
  \begin{subfigure}[t]{0.185\textwidth}
    \includegraphics[width=\linewidth]{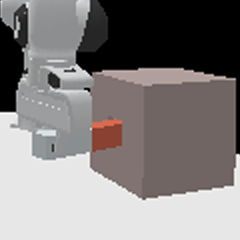}
    \subcaption{RGB image of attempted peg insertion}\label{fig:contact-viz-fig-rgb}
  \end{subfigure}\hfill
  \begin{subfigure}[t]{0.185\textwidth}
    \includegraphics[width=\linewidth]{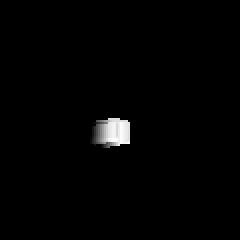}
    \subcaption{Distance to contact on surfaces closest to contact on grasped peg.}\label{fig:contact-viz-fig-ee-dtc}
  \end{subfigure}\hfill
  \begin{subfigure}[t]{0.185\textwidth}
    \includegraphics[width=\linewidth]{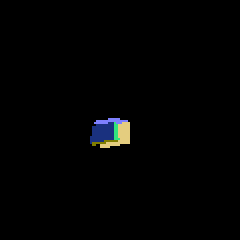}
    \subcaption{Normals of surfaces closest to contact on grasped peg.}\label{fig:contact-viz-fig-ee-normals}
  \end{subfigure}\hfill
  \begin{subfigure}[t]{0.185\textwidth}
    \includegraphics[width=\linewidth]{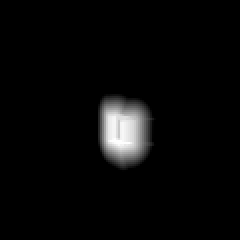}
    \subcaption{Distance to contact on surfaces closest to contact on surrounding environment.}\label{fig:contact-viz-fig-env-dtc}
  \end{subfigure}\hfill
  \begin{subfigure}[t]{0.185\textwidth}
    \includegraphics[width=\linewidth]{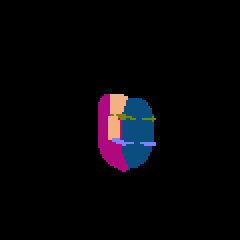}
    \subcaption{Normals of surfaces closest to contact on surrounding environment.}\label{fig:contact-viz-fig-env-normals}
  \end{subfigure}\hfill
  \caption{Visualizations of ground-truth contact-centric features (excluding the contact map) on simulated peg-in-hole task. Fig.~\ref{fig:contact-viz-fig-ee-dtc} and Fig.~\ref{fig:contact-viz-fig-env-dtc} illustrate the distance-to-contact maps for the grasped peg and environment respectively. For ease of visualization, we clamp both to a small value of 5 cm with brightness indicating distance approaches 0. Fig.~\ref{fig:contact-viz-fig-ee-normals} and Fig.~\ref{fig:contact-viz-fig-env-normals} show the normals of surfaces closest to contact (masked within 5 cm) on the grasped peg and environment. In tandem, these features serve to highlight spatial information of surfaces close-to and in-contact, potentially including occluded surfaces}
  \label{fig:contact-viz-fig}
\end{figure*}

We use a diffusion policy architecture~\cite{chiDiffusionPolicyVisuomotor2023a}, employing a ResNet-18 encoder~\cite{heDeepResidualLearning2015a} to condition a U-Net denoiser~\cite{ronnebergerUNetConvolutionalNetworks2015a}. The policy is trained to predict $\textrm{SE}(3)$ end-effector trajectories over a 1.4~s horizon subsampled at 25~Hz from the demonstration data. The diffusion policy operates at 5~Hz, with planned trajectories executed by a low-level impedance controller operating at 1~kHz. To achieve real-time performance on an NVIDIA RTX 2070 SUPER, we tune the denoising diffusion implicit model's scheduler~\cite{songDenoisingDiffusionImplicit2022} to 50 training steps and 10 inference steps. Other training and inference hyperparameters match the original implementation.

Demonstrations are collected via a 3Dconnexion SpaceMouse. The 6D inputs are scaled to a delta $\textrm{SE}(3)$ transformation integrated over time to produce desired end-effector poses at 100~Hz and tracked by the same impedance controller used during policy execution. Recorded data includes 30~Hz RGB-D frames, 1~kHz joint states, 100~Hz desired end-effector poses, and 1~kHz impedance controller states.

To address real-world execution latencies, we linearly interpolate desired end-effector trajectories from the policy by measuring the delay with respect to the timestamp of the policy's image input. In contrast to Chi et al.~\cite{chiUniversalManipulationInterface2024b} where a fixed latency was assumed, by continually measuring delays we can adapt to variability induced by inference times and networking.

The baseline policy, which we refer to as ``training from scratch,'' receives the common standard of current RGB image and end-effector pose as inputs. Extending this: (1) We concatenate the corresponding depth image with the RGB image and (2) include a short history of previous actions, which are encoded using a small 1D convolutional network. Anecdotally, we found these additions improve the data efficiency and performance of the baseline while being straightforward to implement.

We compare the baseline policy against variations augmented with the following pre-trained visual features:
\begin{enumerate}
  \item \textbf{DINOv2} (+DINOv2): A representative task-agnostic latent feature trained via self-supervised learning on real images. Following the recommendations from Theia's authors, we use the model variant incorporating registers~\cite{darcetVisionTransformersNeed2024}, and provide policies with only the dense patch tokens.
  \item \textbf{Theia} (+Theia): A SoTA latent feature tailored for robotics distilling CLIP, DINOv2, and ViT~\cite{shangTheiaDistillingDiverse2024}. The authors claim this combination was best among additional choices of Depth-Anything~\cite{yangDepthAnythingUnleashing2024} and SAM. Theia was also found to outperform robotic-specific VFMs like R3M and MVP.
  \item \textbf{Segmentation Masks} (+Msk): A representative interpretable, object-centric feature. We obtain segmentation masks capturing the grasped object's projected visible geometry. To disambiguate the utility of the representation itself from the quality of estimates, we include in our policy evaluations both estimated masks from SAM-2~\cite{raviSAM2Segment2024} and ground-truth for simulated experiments.
  \item \textbf{Contact-Centric Features} (+Cntct): Another interpretable, object-centric feature inspired by the work of~\cite{yiVisualauditoryExtrinsicContact2025,Kim2023}. Tailored to this study as a proof-of-concept, we attempt to capture spatial information between the grasped object and environment localized to surfaces near and in-contact. Our model ingests a depth image and estimates a probabilistic heatmap of contact regions and for close-to-contact surfaces of the grasped-object and environment: single-channel maps of distances-to-contact and three-channel maps of the normals (Fig.~\ref{fig:contact-viz-fig}). We trained this model via supervised learning using a small-scale synthetic dataset. As with segmentation masks, we compared estimated and ground-truth versions. Additionaly, we find providing a mask of the grasped-object alongside depth improves estimation and include the bolstered model in our evaluations (referred to as ``+Cntct(Msk)'') when possible.
\end{enumerate}

We downsample the raw $480\times640$ RGB-D images to $180\times240$ for each policy. To augment the baseline with pre-trained visual features, we concatenate pre-trained feature maps and RGB-D inputs along the channel dimension when possible. To achieve this for SAM-2 and our contact-centric features, we provide images at $240\times320$ and downsample the outputs to $180\times240$. However, for DINOv2 and Theia's patch features, this yields $15\times20$ patch features that are downsampled to $12\times15$ and instead concatenated between the third and fourth block of the ResNet encoder. While more sophisticated strategies for imparting visual priors may exist, we leave their exploration to future work. In the real world, due to VRAM limits, we evaluate VFMs on a secondary workstation streaming features over UDP.

\section{EXPERIMENTS}

Our experiments evaluate visual representations for visuomotor policy learning in fine-grained, contact-rich tasks, addressing the questions from the introduction. We measure a representation's utility as whether its inclusion improves task success rate over the from-scratch baseline across data budgets. We focus on insertion tasks, which encapsulates the core challenges of this domain: occlusion of critical spatial information, precise positioning demands, and frictional failure modes. We propose two instantiations, each requiring distinct behaviors, and qualify simulation findings with a real-world analogue. Simulated tasks are implemented in ManiSkill~\cite{taoManiSkill3GPUParallelized2025}.

\subsection{Common Setup and Evaluation}
For all tasks, environments feature a single external static RGB-D camera placed across from, and looking at the robot. Honing in on the difficult insertion stage of each task, we initialize the robot with a pre-grasped object. The initial robot pose remains fixed across episodes while object geometries, color, and target locations are randomized, demanding generalization and robustness to irrelevant visuals. We observe that policy performance variance is often dominated by demonstration dataset quality rather than training seeds, especially in lower data regimes. Therefore, for simulated tasks, we train policies at each data budget using five random subsets of demonstrations and report the mean and standard error. Across data budgets, we nest demonstration subsets. i.e. for the same seed, the 30-demonstration subset is fully contained within the 50-demonstration subset. Policies are trained for 300k iterations with checkpoints evaluated every 20k steps on 100 unseen environments to select the model with the highest success rate. Final evaluations are conducted on a unique set of 300 unseen environments. Success is determined by achieving insertion within a specified time budget of $1.2\times$ the duration of the slowest demonstration.

\subsection{Simulated Peg-in-Hole}

\begin{figure}[htb]
    \vspace*{4.5pt}
    \centering
    \begin{subfigure}{0.31\columnwidth}
        \includegraphics[width=\linewidth]{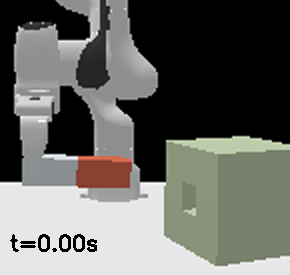}
    \end{subfigure}
    \hfill
    \begin{subfigure}{0.31\columnwidth}
        \includegraphics[width=\linewidth]{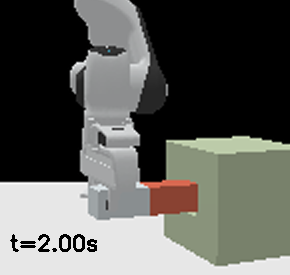}
    \end{subfigure}
    \hfill
    \begin{subfigure}{0.31\columnwidth}
        \includegraphics[width=\linewidth]{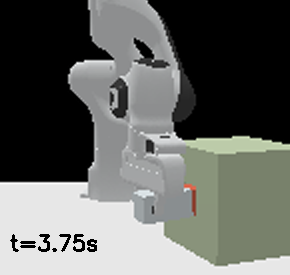}
    \end{subfigure}
    \caption{Simulated peg-in-hole execution at the start, middle, and end. Frames are cropped around the region of interest.}
    \label{fig:sim-peg-in-hole}
\end{figure}

The peg-in-hole task is a canonical problem, known for producing challenging contact phenomena like jamming and wedging~\cite{simunovic1975force,whitneyQuasiStaticAssemblyCompliantly1982a}. We attempt to modulate the severity of such phenomena and demands on precision by evaluating policies across different tolerance values. Along with tight tolerances, a key challenge is that relevant contact surfaces are severely occluded during insertion attempts, as seen in Fig.~\ref{fig:sim-peg-in-hole}.

\paragraph{Randomizations}
The square peg's width and height are sampled uniformly between 1.5~cm and 3~cm, while its length is fixed to 10.5~cm. The hole's color is randomized along with the x-y position by $\pm 5$~cm with a fixed orientation. The hole's dimensions are matched to the peg according to a variable tolerance sampled uniformly between 3~mm and 15~mm during demonstrations. During final evaluations, we investigate the impact of tolerances across three fixed values (3~mm, 9~mm, and 15~mm).

\paragraph{Evaluation}
Final evaluation comprises 300 unseen environments per tolerance level. Success requires inserting at least 5.25~cm of the peg into the hole within 25.86~s.

\subsection{Simulated Book-Insertion}
\begin{figure}[htb]
    \vspace*{4.5pt}
    \centering
    \begin{subfigure}{0.235\columnwidth}
        \includegraphics[width=\linewidth]{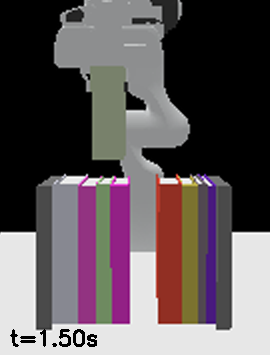}
    \end{subfigure}
    \hfill
    \begin{subfigure}{0.235\columnwidth}
        \includegraphics[width=\linewidth]{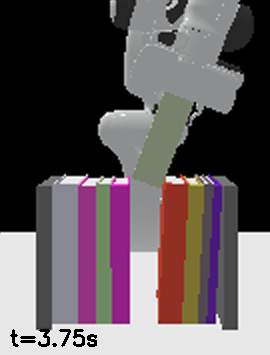}
    \end{subfigure}
    \hfill
    \begin{subfigure}{0.235\columnwidth}
        \includegraphics[width=\linewidth]{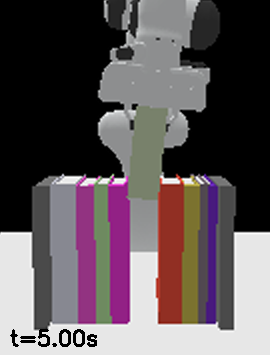}
    \end{subfigure}
    \hfill
    \begin{subfigure}{0.235\columnwidth}
        \includegraphics[width=\linewidth]{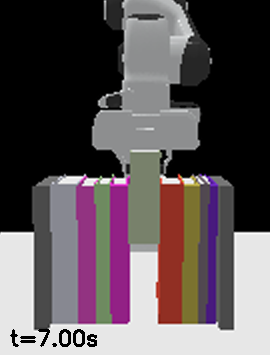}
    \end{subfigure}
    \caption{Simulated book-insertion execution. Frames are cropped around the region of interest.}
    \label{fig:sim-book-insertion}
\end{figure}

\begin{figure}[htb]
    \centering
    
    \begin{subfigure}{0.235\columnwidth}
        \includegraphics[width=\linewidth]{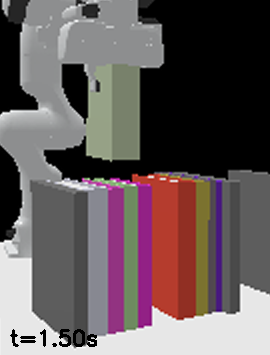}
    \end{subfigure}
    \hfill
    \begin{subfigure}{0.235\columnwidth}
        \includegraphics[width=\linewidth]{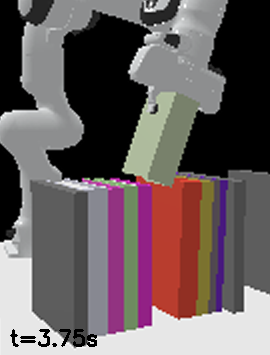}
    \end{subfigure}
    \hfill
    \begin{subfigure}{0.235\columnwidth}
        \includegraphics[width=\linewidth]{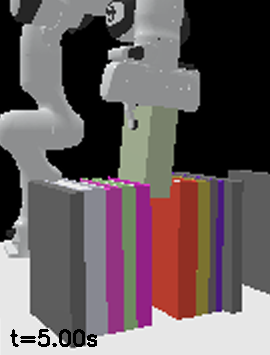}
    \end{subfigure}
    \hfill
    \begin{subfigure}{0.235\columnwidth}
        \includegraphics[width=\linewidth]{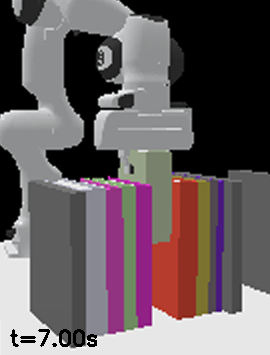}
    \end{subfigure}
    
    \vspace{1em} %
    
    \begin{subfigure}{0.235\columnwidth}
        \includegraphics[width=\linewidth]{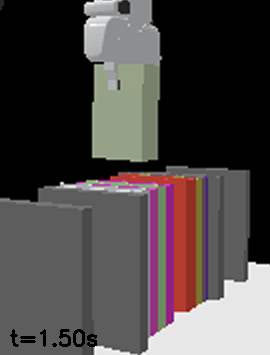}
    \end{subfigure}
    \hfill
    \begin{subfigure}{0.235\columnwidth}
        \includegraphics[width=\linewidth]{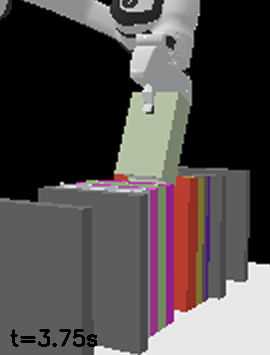}
    \end{subfigure}
    \hfill
    \begin{subfigure}{0.235\columnwidth}
        \includegraphics[width=\linewidth]{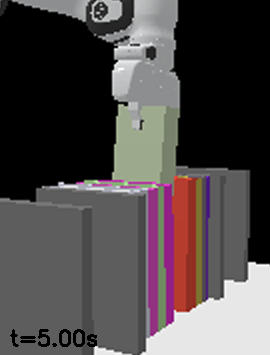}
    \end{subfigure}
    \hfill
    \begin{subfigure}{0.235\columnwidth}
        \includegraphics[width=\linewidth]{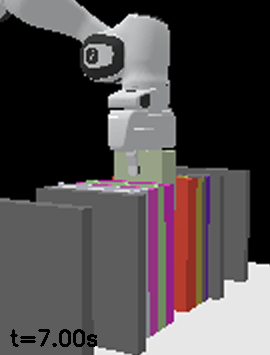}
    \end{subfigure}
    
    \caption{Book-insertion task at re-rendered views. First row shows camera pivoted by \ang{30} while second shows \ang{60}. Frames are cropped and centered to capture the key region of interest.}
    \label{fig:sim-book-insertion-views}
\end{figure}

Unlike the rigidly constrained peg-in-hole task, book-insertion involves dynamic shelved books supported by spring-loaded prismatic bookends, approximating the compliance of commonly found metal ones. However, the slot is intentionally too narrow by 3.5~mm, requiring the robot to leverage forceful interactions to pry adjacent books apart and establish sufficient clearance before proceeding with insertion (Fig.~\ref{fig:sim-book-insertion}). While the nominal viewpoint ensures grasped book and slot boundaries remain visible throughout the task, we investigate the impact of task-relevant occlusions by pivoting the camera viewpoint about the scene by \ang{30} and \ang{60} (Fig.~\ref{fig:sim-book-insertion-views}), while holding the number of training demonstrations fixed at 30. To maintain consistent training data across views, we play back recorded states and re-render images and their corresponding pre-trained visual features.

\paragraph{Randomizations}
The grasped book's width is uniformly sampled between 3~cm and 6.5~cm, and its height between 16.5~cm and 25~cm. Its color is also randomized. To simplify the task, the surrounding shelved books are fixed in shape and color. The target slot location is randomized laterally by $\pm 5$~cm.

\paragraph{Evaluation}
Final evaluation comprises 300 unseen environments per viewpoint. Success requires inserting the book with at most 7~cm of vertical overhang within 12.66~s.

\subsection{Real-World Book-Insertion}
\begin{figure}
    \vspace*{4.5pt}
    \centering
    \includegraphics[width=0.6\linewidth]{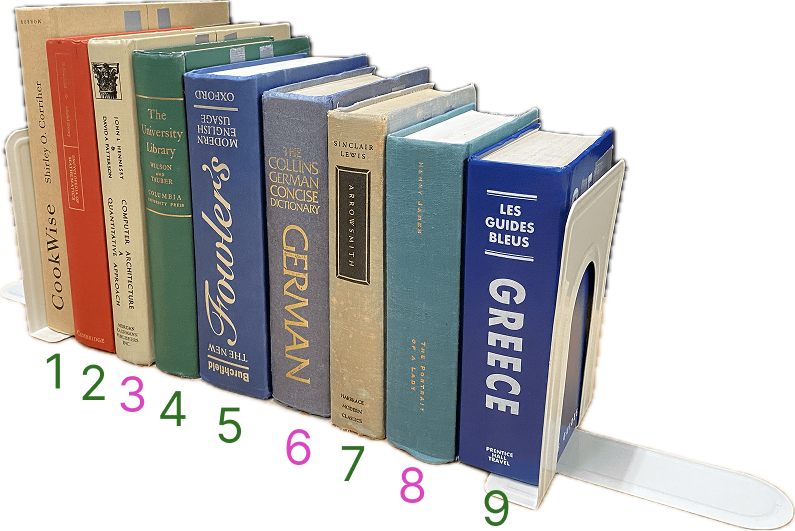}
    \caption{Books used for real-world insertion. Green denotes the six training books; pink denotes the three evaluation books. The same metal bookends support the shelved books.}
    \label{fig:real-world-grasped-books}
\end{figure}

To verify our findings from simulation, we constructed a real-world analogue of the simulated book-insertion task (Fig.~\ref{fig:teaser}). We match the simulated environment closely, imitating the nominal camera viewpoint, shelf dimensions, and spring-like compliant metal bookends. 

\paragraph{Randomizations}
We use 6 grasped books with varied dimensions and colors for training (Fig.~\ref{fig:real-world-grasped-books}), collecting 48 demonstrations per book for a total of 288. Unlike simulation, we fix the slot location to ease data collection.

\paragraph{Evaluation}
We train one policy per approach for a fixed 300k iterations across two data budgets: the full 288 demonstrations and a reduced set of 192 (holding out two books). Final evaluation consists of 10 attempts on each of three unseen books, across all data budgets. We report 95\% binomial confidence interval bounds for these 30 evaluations using the technique described by Vincent et al.~\cite{vincentHowGeneralizableMy2024}. The success time budget is 18.5~s.

\section{RESULTS}

\begin{figure*}[ht] 
  \centering
  \begin{subfigure}[t]{0.35\textwidth}
    \includegraphics[width=\linewidth]{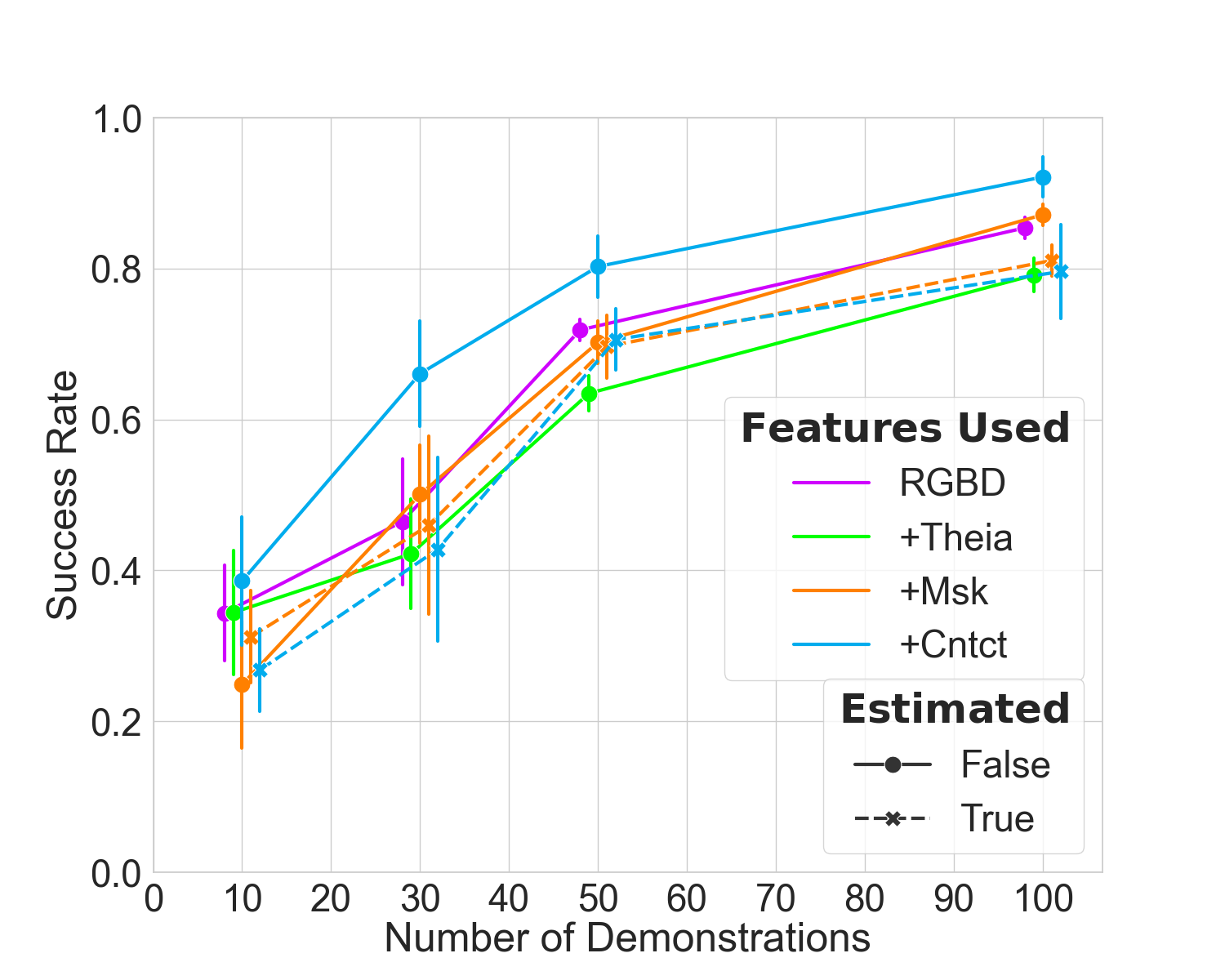}
    \subcaption{Evaluations at 15~mm tolerance}\label{fig:simulated-peg-insertion-evals-15mm}
  \end{subfigure}\hfill
  \begin{subfigure}[t]{0.30\textwidth}
    \includegraphics[width=\linewidth]{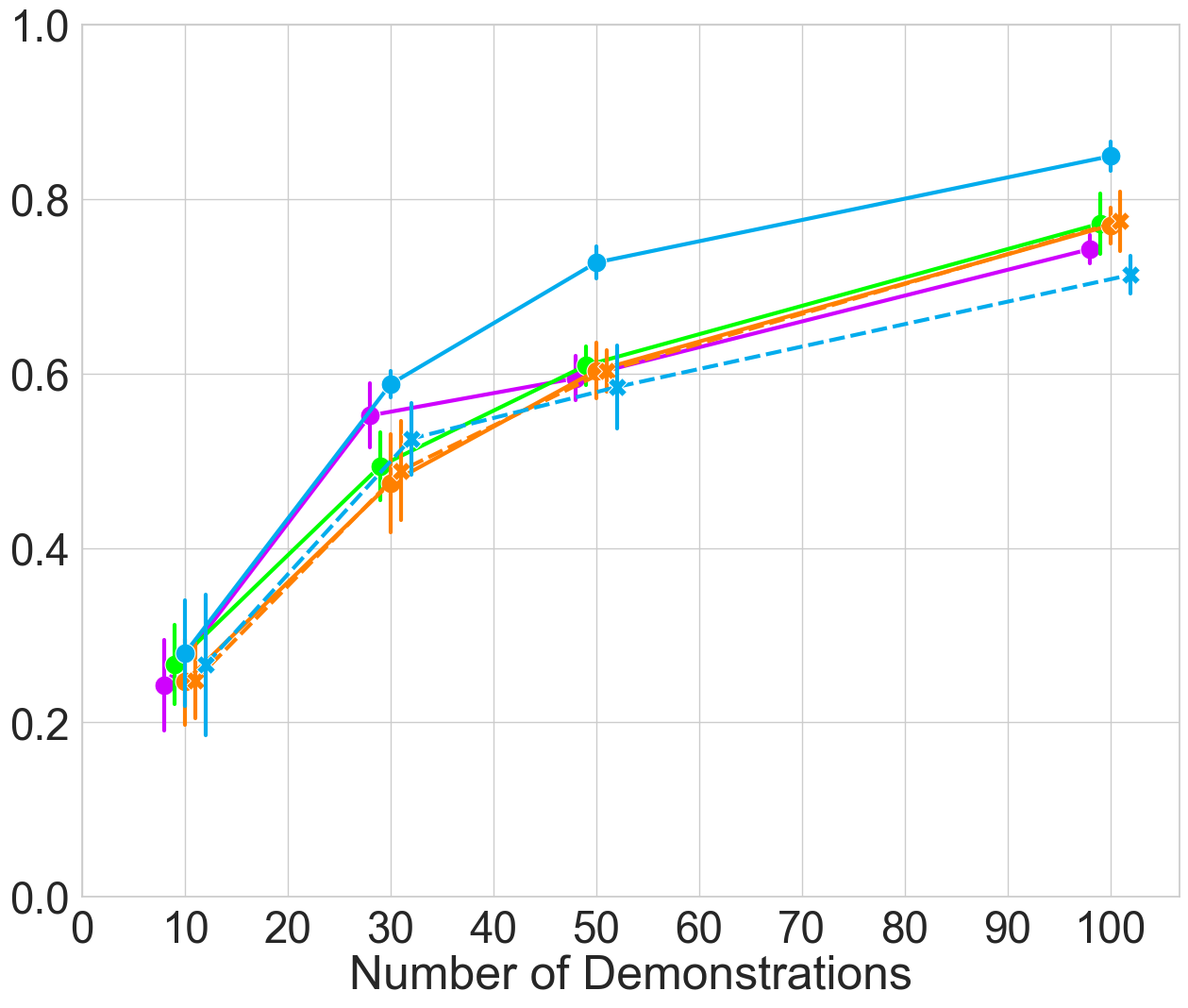}
    \subcaption{Evaluations at 9~mm tolerance}\label{fig:simulated-peg-insertion-evals-9mm}
  \end{subfigure}\hfill
  \begin{subfigure}[t]{0.30\textwidth}
    \includegraphics[width=\linewidth]{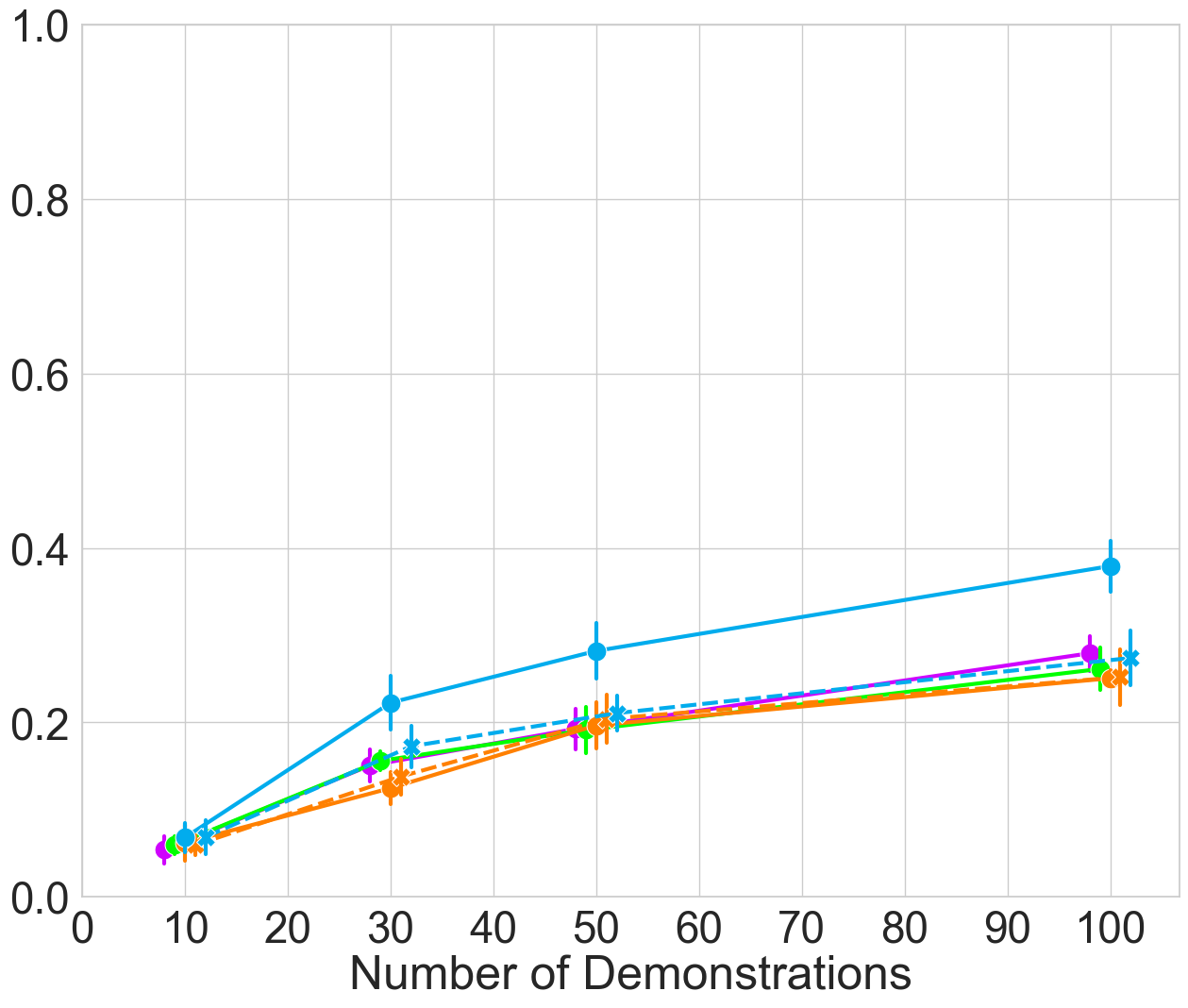}
    \subcaption{Evaluations at 3~mm tolerance}\label{fig:simulated-peg-insertion-evals-3mm}
  \end{subfigure}\hfill

  \captionof{figure}{Results for the simulated peg-in-hole task across three fixed tolerances. Error bars indicate standard error across 5 randomized subsets of demonstrations.}
  \label{fig:simulated-peg-insertion-evals}
\end{figure*}

\begin{figure}[t]
    \centering
    \includegraphics[width=0.95\linewidth]{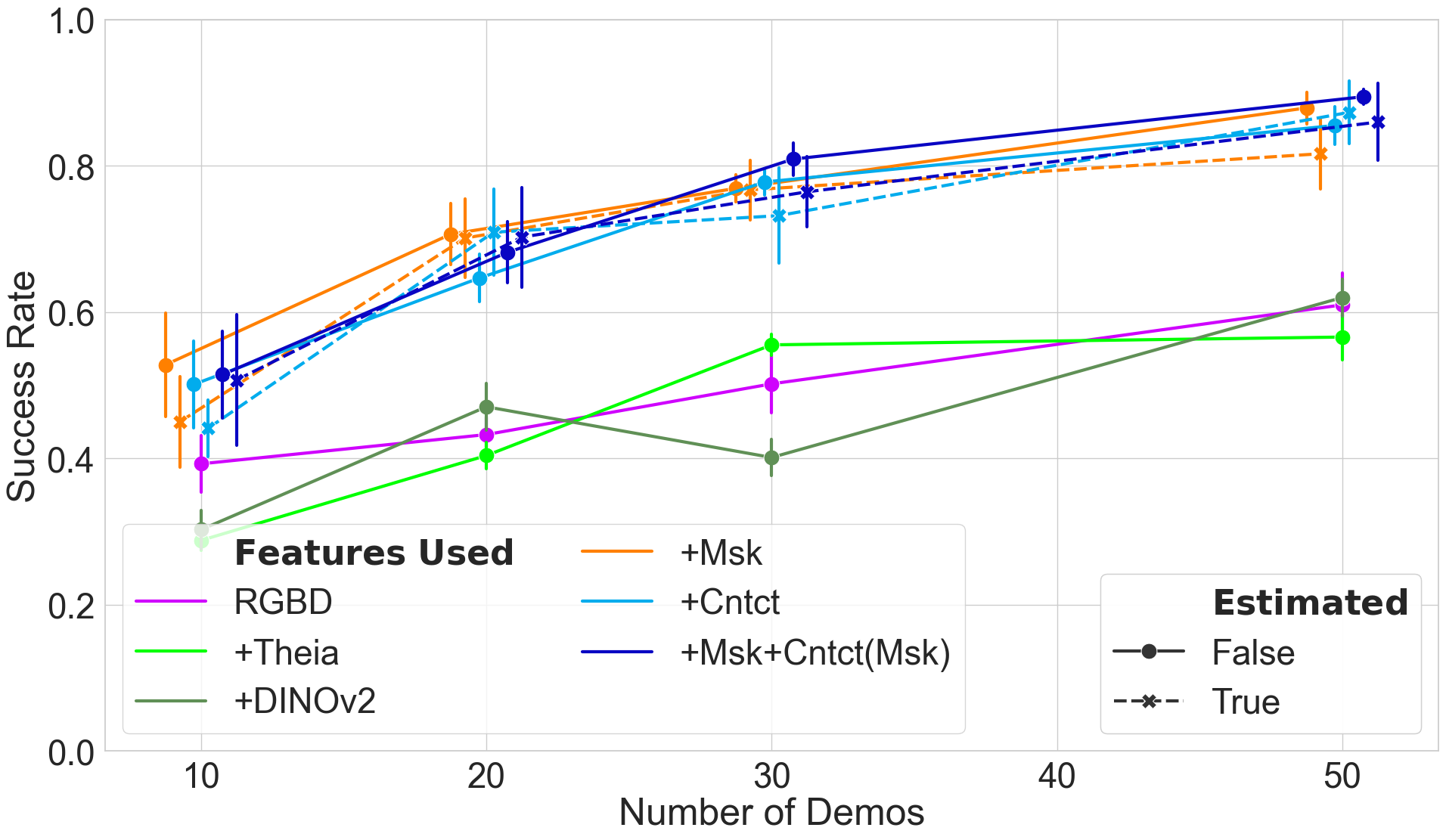} 
    \caption{Results for the simulated book-insertion task. Error bars indicate standard error across 5 randomized subsets of demonstrations.}
    \label{fig:simulated-book-insertion-evals}
\end{figure}

\begin{figure}[t]
    \centering
    \includegraphics[width=0.95\linewidth]{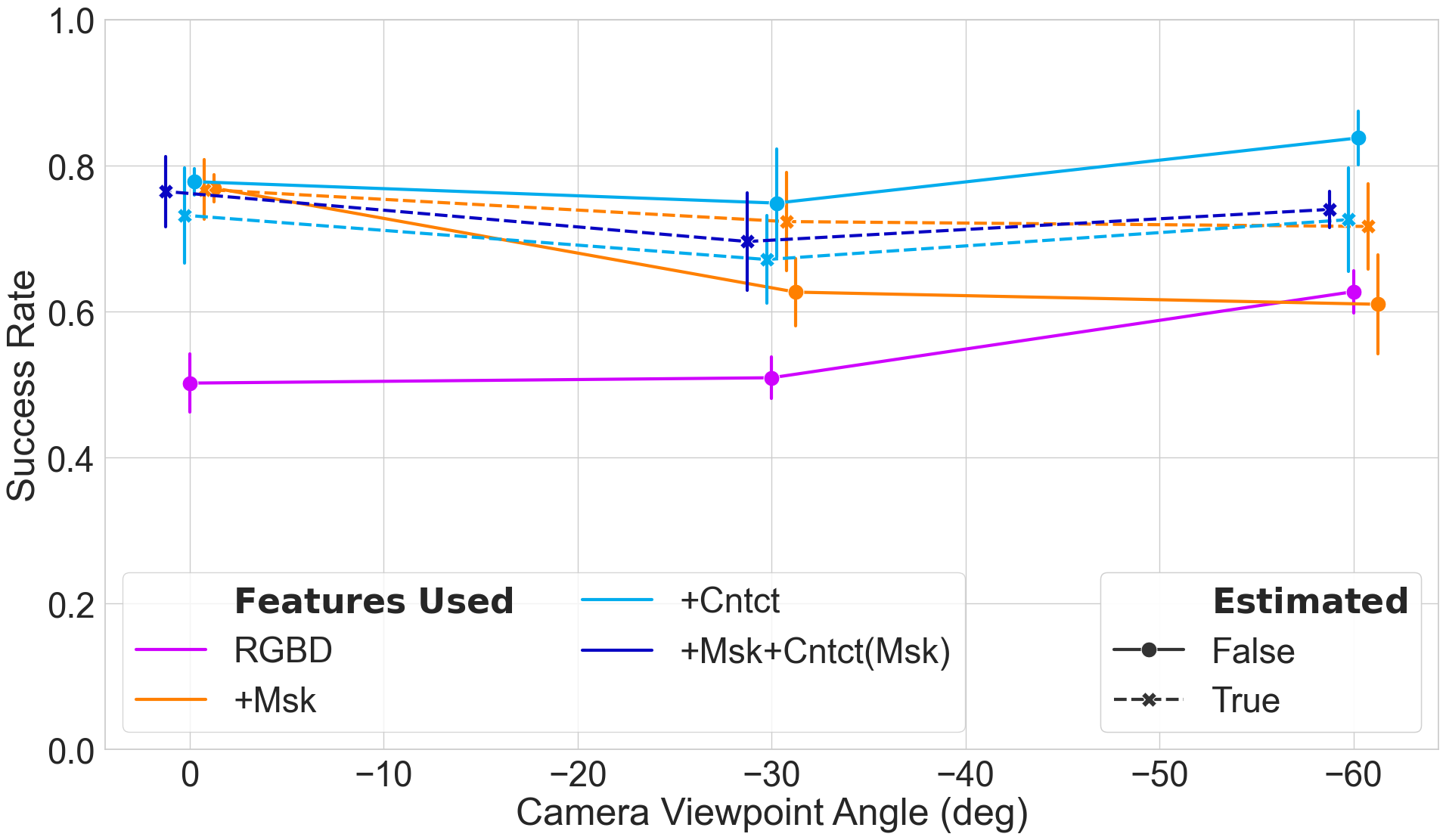} 
    \caption{Results for simulated book-insertion using 30 demonstrations across different viewpoints. Error bars indicate standard error across 5 randomized subsets of demonstrations.}
    \label{fig:simulated-book-insertion-views-evals}
\end{figure}

\begin{figure}[t]
    \centering
    \includegraphics[width=0.95\linewidth]{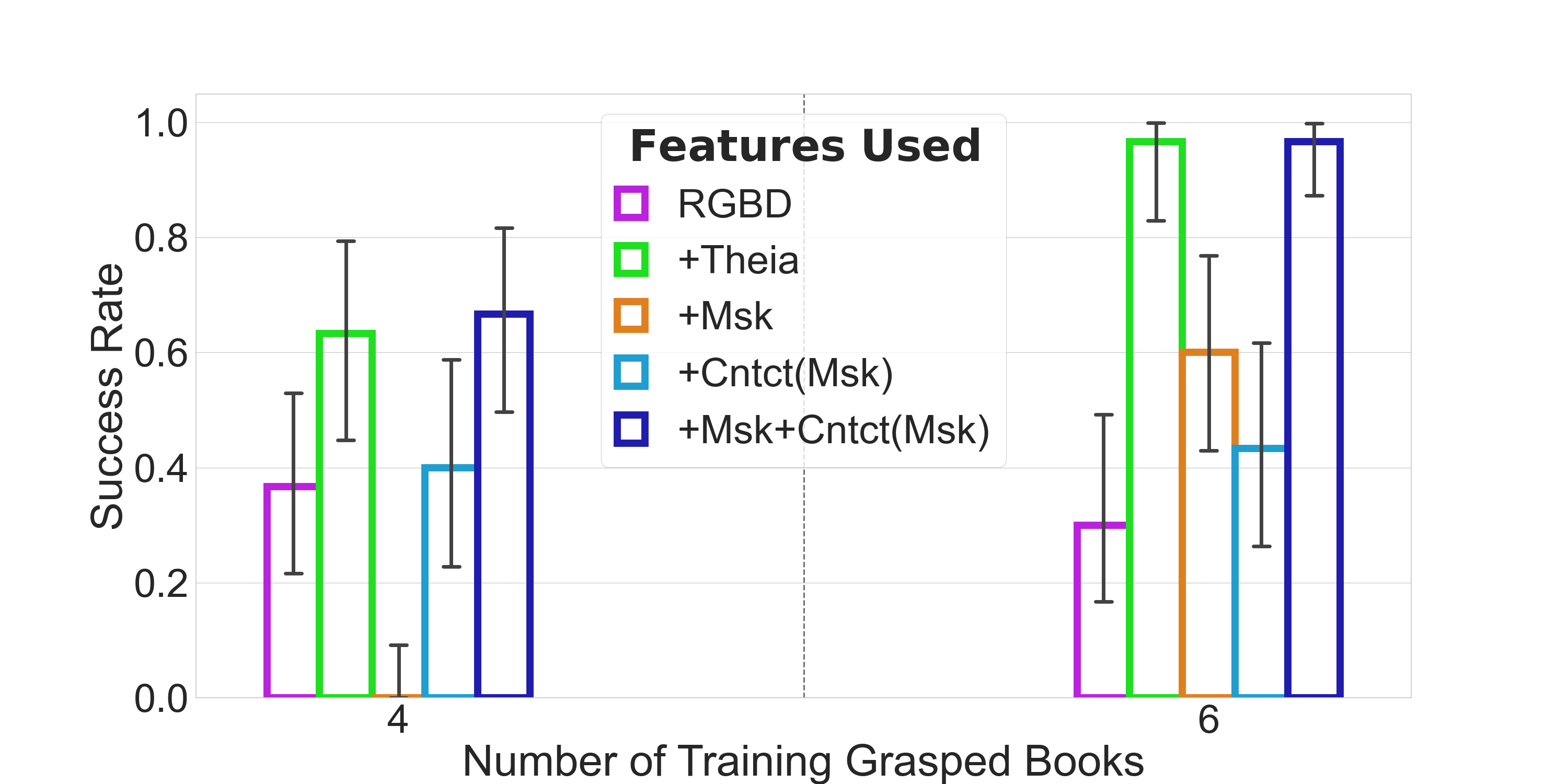} 
    \caption{Results for the real-world book-insertion task. Error bars indicate bounds at 95\% confidence intervals across 30 evaluation episodes.}
    \label{fig:real-world-book-insertion-evals}
\end{figure}

Our results suggest that the impact of visual representations are highly task-dependent in our setting and often challenged our initial expectations. We organize our analysis around the motivating questions, which we find most salient for understanding the results and guiding future research directions. While we conducted extensive evaluations in simulation, practical considerations make this difficult in the real world. Nevertheless, we observe several trends that raise interesting questions.

\subsection{Common General-purpose VFMs}
Despite the state-of-the-art performance reported by Theia across simulated benchmarks, both it and DINOv2 failed to outperform the from-scratch baseline in the simulated book-insertion task (Fig.~\ref{fig:simulated-book-insertion-evals}). This suggests that neither DINOv2 on its own or the added distillations of CLIP and ViT provided by Theia's features confers significant benefits in this setting. We note these results do not necessarily contradict the findings from Theia's authors where out of 12 simulated tasks, Theia performed significantly better than DINOv2 on only 5. Due to its state-of-the-art status and robotics-tailored design, we maintain Theia as a representative general-purpose VFM for the remaining tasks. Results in the simulated peg-in-hole setting echo those from book-insertion, with Theia performing on par with the baseline across tolerances and degrading similarly as tolerances are tightened (Fig.~\ref{fig:simulated-peg-insertion-evals}).

Though Theia failed to outperform the baseline in simulation, we find it provides significant boosts in the real-world book-insertion task at both low and high data budgets (Fig.~\ref{fig:real-world-book-insertion-evals}). A plausible explanation is real-to-sim distribution shift: Theia is distilled using real images from ImageNet~\cite{dengImageNetLargescaleHierarchical2009} and may be negatively impacted when applied to synthetic data. This raises questions about the widespread use of simulation to benchmark representations trained on real images~\cite{majumdarWhereAreWe2024,shangTheiaDistillingDiverse2024}, and whether such evaluations may be misleading for real-world performance.

Meanwhile, segmentation masks in both estimated and ground-truth variants provide substantial gains in the simulated book-insertion task over both general-purpose VFMs and training from scratch (Fig.~\ref{fig:simulated-book-insertion-evals}). Despite distilling several state-of-the-art priors, that Theia underperforms compared to masks suggests capturing explicit object geometry may be of greater importance in this setting. However, these gains are not reflected in the peg-in-hole setting (Fig.~\ref{fig:simulated-peg-insertion-evals}) where, like Theia, performance closely tracks with the baseline across tolerances. This suggests that while a simple object-centric representation of the grasped object via its projected surfaces can be sufficient for tasks where relevant object boundaries are visible, masks provide limited value when such boundaries and the object itself become occluded. 

To test this, we perturbed the camera extrinsics of the book-insertion task towards increasingly oblique views of the slot. We anticipated that this would induce occlusions of task-relevant surfaces and hence attenuate the gains of segmentation masks as seen in the peg-in-hole setting. However, we found only ground-truth masks suffered significantly while other approaches largely maintained their performance across views (Fig.~\ref{fig:simulated-book-insertion-views-evals}). We suspect the surrounding books, which react to the insertion process, may have provided compensatory visual cues (Fig~\ref{fig:sim-book-insertion-views}), whereas the ground-truth mask-based policy may have been overly reliant on visibility of the grasped object. Noise in the estimated masks may have dampened this sensitivity, allowing it to maintain an edge over the baseline, and resulting in the unintuitive gain from ground-truth to estimated masks. 

Real world evaluations of the book-insertion task partially support the gain of segmentation masks (Fig.~\ref{fig:real-world-book-insertion-evals}), where masks achieves a significant boost over training from scratch at the higher data budget, but not the lower one. Although similar considerations of real-to-sim distribution shift apply to SAM-2, unlike latent VFMs, the interpretability of SAM-2's explicit outputs allowed us to qualitatively verify feature consistency between simulation and the real world.

\subsection{Contact-Centric Features}

Ground-truth contact-centric information is consistently among the best representations across all simulated tasks and is the only representation to significantly outperform the baseline across all peg-insertion tolerances. These margins are most prominent in the low and mid data regimes (30 and 50 demonstrations) and at tighter tolerances (3~mm and 9~mm). However, it is closely matched by segmentation masks in the simulated book-insertion setting (Fig.~\ref{fig:simulated-book-insertion-evals}). This suggests that explicitly encoding the geometric relationship between the grasped object and environment, especially of surfaces near and in-contact, is especially useful in tightly-toleranced insertion tasks. With the caveat that when relevant interfaces are clearly visible, other spatially-grounded representations such as segmentation masks may provide sufficient structure for inferring task-critical information. 

Similarly, we found that the estimated contact-centric features are as good as their ground-truth counterparts for book-insertion (Fig.~\ref{fig:simulated-book-insertion-evals}), with both largely maintaining their edge over the baseline across changing viewpoints (Fig.~\ref{fig:simulated-book-insertion-views-evals}). This seems to be anecdotally reflected in the estimation quality of our contact features which, while noticeably degraded, still manage reasonable contact-region estimates despite greater occlusions. However, these gains failed to transfer to the peg-in-hole task (Fig.~\ref{fig:simulated-peg-insertion-evals}). This likely reflects the sustained occlusion of critical faces during insertion attempts of the peg, making inference of relative configurations and contact difficult. Moreover, in the real-world book-insertion task, estimated contact-centric features do not provide significant gains over training from scratch at either data budget, likely due to the challenges of accurate estimation in the presence of real-world sensor noise. However, the combination of contact-centric features and masks yields a significant boost across data budgets, outperforming either alone and the baseline, suggesting these two representations may provide synergistic benefits in real-world settings.

\section{DISCUSSION AND CONCLUSION}

We present a controlled study of visual representations for visuomotor policy learning in fine-grained, contact-rich manipulation. Rather than surveying a broad task distribution, we focus explicitly on two closely-related insertion tasks, a rigidly constrained peg-in-hole setup and a dynamic book-insertion scenario, and perturb each with targeted variations in physical tolerance and task-relevant occlusion. Across these settings, we compare canonical representations from vision foundation models against a hypothesis-driven representation that explicitly encodes contact-centric information.

Our results underscore the highly task-dependent and often unpredictable nature of which visual priors are most useful for policy learning. Relative trends differ drastically across settings and even fluctuate unexpectedly over subtle variations within the same task. Task dynamics, viewpoints, sensor noise, and visual clutter all seem to modulate which priors provide value. These results echo community-wide findings, but we believe that the contact-rich domain, with stiff, complex dynamics and occlusions, is especially resistant to one-size-fits-all strategies. Instead, representation choice in this domain currently demands per-task empirical validation, reinforcing our claim that no reliable prescription yet exists.

In spite of this variability, one result held across every setting: ground-truth contact information was the only representation to deliver consistent gains, underscoring the importance of directly capturing geometric relationships between the manipuland and its environment. Our proof-of-concept contact-centric representation shows that such features can improve learning outcomes when estimates are accurate. By contrast, widely-utilized VFM features like DINOv2 and Theia failed to outperform from-scratch training in simulation experiments. When task-relevant occlusions are minimal, segmentation masks offer an off-the-shelf solution that confers comparable gains to contact-centric features. However, under realistic visual complexity and sensor noise, accurate spatial relationships and contact cues are difficult to infer. In these settings, combining segmentation masks with contact-centric features seems to yield synergistic benefits, outperforming either alone. These findings motivate advancing contact perception capabilities and their integration into visuomotor policy learning. While dense proximity or tactile sensing at key interfaces directly captures critical information, such instrumentation is often impractical. A pragmatic path may be to fuse strong shape priors and pose estimation from vision with distal haptic or tactile sensing to infer the inter-object states that govern contact mode transitions.

Several factors qualify our conclusions. First, variance across randomized demonstration subsets suggests demonstration quality is a major driver of policy performance. Furthermore, we note that this intra-seed variance may also be attributable to checkpoint evaluation frequency. Future work should disentangle these effects to clarify their roles and interactions. We also note some limitations. Confidence in real-world results is bottlenecked by practical constraints, such as early stopping, reproducibility of environment conditions across policy rollouts, and limited evaluation scale. Higher-fidelity simulation and sophisticated statistical methods may offer a tractable alternative to scaling real-world trials~\cite{teamCarefulExaminationLarge2025a}. Finally, extending this controlled, magnifying-glass methodology to other contact-rich settings, especially those demanding delicate yet forceful interaction, would test whether the instability we document is endemic to the domain, and remains an important avenue for future work

\section*{ACKNOWLEDGMENTS}
\ifanonymized
\else
Leon Kim was supported by the NSF Graduate Research Fellowship Program. Dinesh Jayaraman was supported by NSF CAREER 2239301, DARPA TIAMAT HR0011249042, ONR N00014-22-1-2677, and NSF SLES 2331783. Michael Posa was supported by an NSF CAREER Award under Grant No. FRR-2238480 and the RAI Institute.\fi

\bibliographystyle{IEEEtran}
\bibliography{./common/extrinsic_contact,./common/special_command}

\end{document}